\pdfoutput=1
\documentclass[11pt]{article}
\usepackage{authblk}

\usepackage{arxiv}

\usepackage{times}
\usepackage{latexsym}

\usepackage[T1]{fontenc}
\usepackage[utf8]{inputenc}

\usepackage{microtype}

\usepackage{inconsolata}

\usepackage{amssymb}        % for star next to GMP
\usepackage{makecell}       % for multi-row cells in tables
\usepackage{multirow}
\usepackage[pdftex]{graphicx} % suggested by NeurIPS22 template
\usepackage{caption}  % for subfigure
\usepackage{subcaption} % for subfigure
\usepackage{booktabs}       % professional-quality tables

\newcommand{\bert}{$\textrm{BERT}_{\tiny{\textrm{BASE}}}\,$}
\newcommand{\bertL}{$\textrm{BERT}_{\tiny{\textrm{LARGE}}}\,$}
\newcommand{\gmp}{GMP$\scriptstyle\bigstar$\,}
\newcommand{\gmpmvp}{GMP$_{\small{\textrm{MvP}}}$\,}
\newcommand{\gmplth}{GMP$_{\small{\textrm{LTH}}}$\,}
\newcommand{\gmpofa}{GMP$_{\small{\textrm{Prune OFA}}}$\,}
\newcommand{\obert}{oBERT$\scriptstyle\bigstar$\,}

\newcommand{\comment}[1]{}

\title{\gmp: Well-Tuned Gradual Magnitude Pruning Can Outperform Most BERT-Pruning Methods}

\begin{document}

\author[1]{Eldar Kurtic\thanks{~~~Corresponding author: eldar.kurtic@ist.ac.at.}~~}
\author[1,2]{Dan Alistarh}
\affil[1]{Institute of Science and Technology Austria}
\affil[2]{Neural Magic Inc.}

\maketitle
\begin{abstract}
  We revisit the performance of the classic gradual magnitude pruning (GMP) baseline for large language models, focusing on the classic BERT benchmark on various popular tasks. Despite existing evidence in the literature that GMP performs poorly, we show that a simple and general variant, which we call \gmp, can match and sometimes outperform more complex state-of-the-art methods. 
  Our results provide a simple yet strong baseline for future work, highlight the importance of parameter tuning for baselines, and even improve the performance of the state-of-the-art second-order pruning method in this setting.
\end{abstract}

\section{Introduction}
The massive recent growth of the computational cost of accurate deep learning models, in particular large language models (LLMs), has motivated the development of several advanced {model compression} techniques~\citep{hoefler2021sparsity, gholami2021survey}, encompassing unstructured and structured pruning, quantization, and knowledge distillation. 
In this paper, we focus on the unstructured pruning, for which we follow the standard pipeline. Such models are first \emph{pre-trained} on a large \emph{upstream} corpus of unlabelled text. Then, they are \emph{fine-tuned} in a supervised manner on a smaller \emph{downstream} task, such as question-answering or text classification.  
In the context of compression, this pipeline led to two paradigms: 1) \emph{upstream pruning}, followed by fine-tuning of the remaining weights on a downstream task, and 2) \emph{downstream pruning}, pruning and fine-tuning directly on the downstream task.

A tempting baseline approach in most settings is \emph{gradual magnitude pruning (GMP)}~\citep{hagiwara1994, zhu2017prune}, that is, periodically removing the smallest fraction of weights during training, possibly interspersed with fine-tuning steps designed to recover accuracy. 
GMP has been shown to be an extremely strong baseline in the context of computer vision~\citep{gale2019state, hoefler2021sparsity}.
However, the literature on pruning LLMs, and in particular BERT models~\cite{sanh2020movement, chen2020lottery, zafrir2021prune}, clearly suggests that GMP \emph{does not} perform well. 

\begin{figure}
    \centering
    \includegraphics[width=\linewidth]{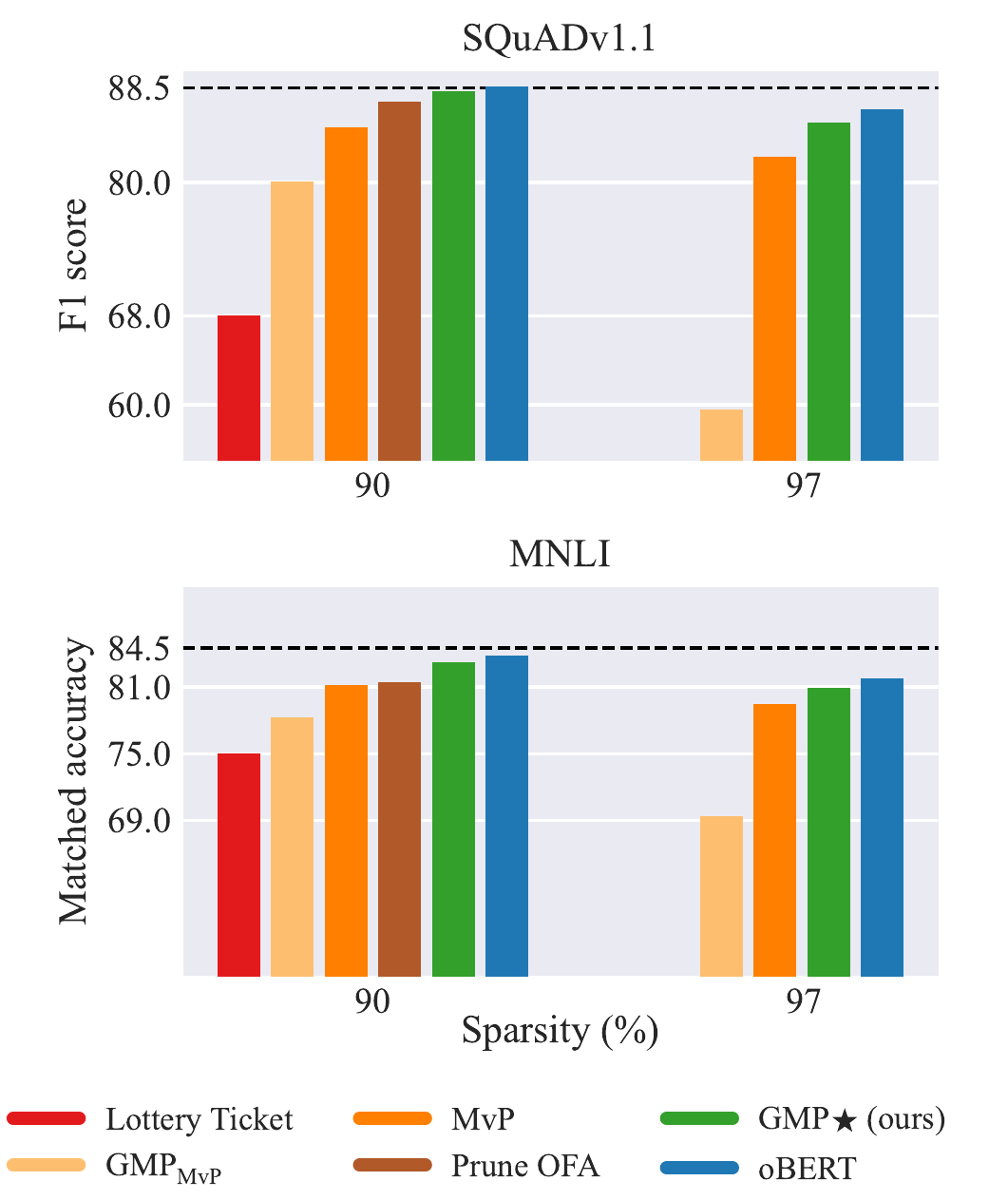}
    \caption{Performance of state-of-the-art unstructured pruning methods relative to the dense \bert model at high sparsities and two tasks, SQuADv1.1 and MNLI.}
    \label{fig:overview}
    \vspace{-0.2in}
\end{figure}

\paragraph{Contribution.} In this paper, we re-examine this conclusion and investigate whether GMP can be a competitive baseline, once carefully tuned. 
Specifically, we show that a well tuned variant which we call \gmp, can produce highly accurate and sparse language models in both upstream and downstream pruning regimes, matching or even  outperforming more complex methods. We explore effects of the crucial parameters for gradual pruning, and provide simple and intuitive guidelines on how to integrate them in a principled manner.  

Our results are summarized in Figure~\ref{fig:overview}, which presents performance of state-of-the-art unstructured pruning techniques on two benchmarks. Specifically, we compare \gmp with the Lottery Ticket approach~\citep{chen2020lottery}, Movement Pruning (MvP)~\citep{sanh2020movement} (as well as its GMP baseline $\textnormal{GMP}_{\textnormal{MvP}}$), upstream Prune OFA~\citep{zafrir2021prune}, as well as the recently-proposed second-order pruning oBERT~\citep{kurtic2022optimal}. 
We observe that: 1) for both benchmarks, \gmp is only second to the more complex oBERT method; 2) \gmp in fact outperforms the highly competitive Prune OFA and MvP methods; and 3) \gmp outperforms both Lottery Tickets and $\textnormal{GMP}_{\textnormal{MvP}}$ by extremely wide margins. 

\comment{%this paragraph seems a bit lame, we are not Greta
%Given that almost all methods, more sophisticated than the data-free magnitude pruning, rely on higher-order information like gradients \cite{sanh2020movement, lagunas2021block}, loss curvature \cite{singh2020woodfisher, frantar2021m, kurtic2022optimal}, and pre-training setups that are expensive to run \cite{chen2020lottery, zafrir2021prune}, the pool of researchers and practitioners who are able to apply them may be limited. We hope that our competitive GMP setup will enable more people to produce sparse and accurate LLMs in compute-limited environments, as well as to provide strong baselines for the development of better pruning techniques.
}

\paragraph{Prior Work.} 
Following the vast BERT-pruning literature, we focus on the unstructured pruning of the \bert model~\citep{devlin2018bert}. As previously noted, upstream and downstream pruning paradigms exist, and methods are usually developed and specialized for only one of the two. For example, Movement Pruning (MvP)~\citep{sanh2020movement, lagunas2021block} for downstream pruning and Prune Once for All (Prune OFA)~\citep{zafrir2021prune} for upstream pruning. Simplicity and generality of the GMP makes it suitable for both paradigms, without any regime-specific modifications. New and more advanced pruning techniques, which are, contrary to GMP, able to leverage gradients~\citep{sanh2020movement, lagunas2021block}, loss curvature~\citep{kurtic2022optimal}, compute-intensive pre-training setup~\citep{zafrir2021prune} are built on the premise that the simple magnitude-based GMP method falters when applied to BERT-pruning. In this work, contrary to what is currently available in the literature, we present empirical evidence that GMP, when tuned carefully, can produce very accurate sparse models which are competitive or even better than most state-of-the-art pruning techniques across both regimes (upstream and downstream). As can be seen from Figure \ref{fig:overview} and our later results, we massively improve upon existing GMP-based pruning baselines, in some cases by even more than \textbf{20 accuracy points}.

\section{Competitive Gradual Magnitude Pruning (\gmp)}
\paragraph{Experimental setup.}
% For a given model $\w \in \mathbb{R}^d$, we seek to identify a subset of weights whose pruning (setting to zero) won't harm model's performance. This is usually done in a gradual setup where number of pruned weights (sparsity) increases over time. Magnitude pruner identifies these less important weights by assigning a score to each weight $w_i$ based on its magnitude (absolute value) $\rho_i = | w_i |$. As the common wisdom suggests, the ones with the lowest score contribute the least and could be safely removed.
We focus our attention on the standard \bert model, composed of embedding and encoder layers, which has approximately 110M parameters. All methods focus on pruning among approximately 85M weights of encoder layers and report sparsities with respect to that number. We evaluate models on the validation split of the respective dataset, and to improve confidence in the obtained results we perform multiple runs with different seeds and report mean performance.
\subsection{Downstream pruning}
Following the literature, we consider three popular tasks: question-answering SQuADv1.1~\citep{rajpurkar2016squad}, recognition of textual entailment MNLI ~\citep{williams2017broad}, and duplicate question detection QQP~\citep{iyer2017first}. Now, we reflect upon the most important constituents of the gradual pruning framework that enabled us to attain massive improvements.

\begin{figure}[t]
    \centering
    \includegraphics[width=\linewidth]{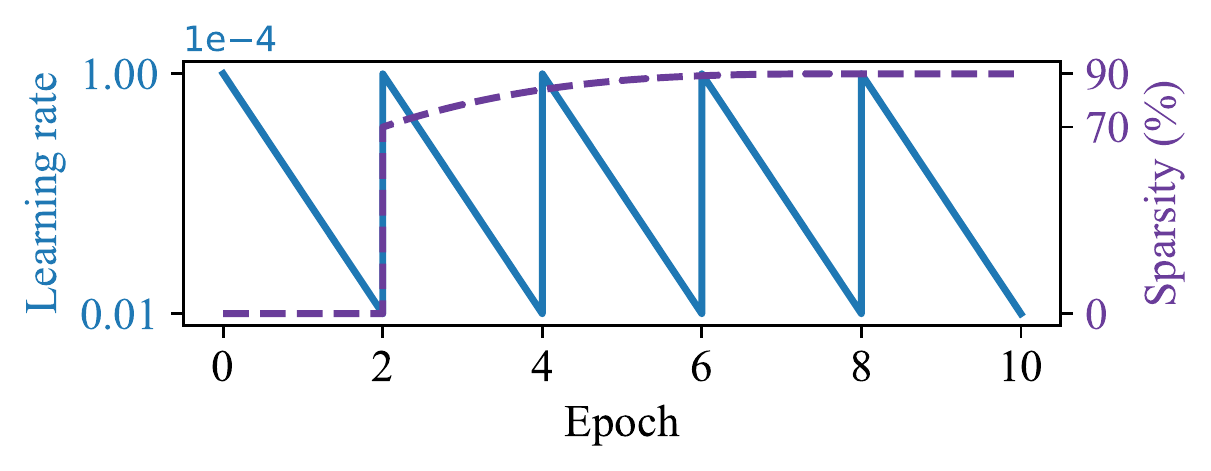}
    \vspace{-0.25in}
    \caption{Learning rate and sparsity schedules for the proposed gradual pruning framework.}
    \label{fig:lr_and_spars}
    \vspace{-0.2in}
\end{figure}

\paragraph{Sparsity schedule.}
In all of our gradual runs, there is no pruning during the first two and the last two epochs. The former fine-tunes the pre-trained model, and the latter fine-tunes the sparse model with the fixed mask. In between the two, \gmp follows the cubic sparsity scheduler~\citep{zhu2017prune} and prunes weights with the frequency of ten times per epoch. Motivated by the fact that \bert is heavily overparametrized for downstream tasks, we deviate from the standard cubic schedule by introducing a large first pruning step. This showed to be of a crucial importance when pruning the model to high target sparsities (e.g. 97\%) as it leaves more time to recover from later pruning steps which are much more difficult. In Table~\ref{tab:initsparsity_sweep} we report results from an ablation study with respect to the size of the initial step. For convenience, we visualize the sparsity scheduler in Figure~\ref{fig:lr_and_spars}. Our preliminary experiments showed similar performance between uniform and global sparsity distributions, so we use the former. 

\paragraph{Learning rate schedule.} 
Our goal is to provide a simple baseline setup that works well across wide range of datasets without any additional task-dependent tuning. Currently, papers either report best results following an extensive hyperparameter search for each task, e.g.~\citet{zafrir2021prune}, or they make use of carefully crafted schedulers for each setup independently which may include warm-up phases with and without rewinds~\citep{sanh2020movement, kurtic2022optimal}. This may lead to high specialization to the target task/model, which is undesirable in practice and makes it hard to distinguish benefits from the pruning technique itself. We propose to simply \textit{replicate} the standard 2-epoch fine-tuning schedule~\citep{devlin2018bert} by a certain factor and intertwine it with pruning steps. For a fair comparison with~\citet{sanh2020movement} we replicate it by a factor of 5, reproducing their 10-epoch setup. And for a fair comparison with~\citet{chen2020lottery} we replicate it by a factor of 15, reproducing their 30-epoch setup. For convenience, we visualize the learning rate schedule in Figure~\ref{fig:lr_and_spars}. In appendix~\ref{app:failed_lr}, we describe results with other schedulers that didn't work.

\paragraph{Knowledge Distillation (KD) Hardness.} We leverage KD~\cite{hinton2015distilling} of outputs from a fine-tuned dense teacher. KD is a standard practice when pruning, e.g.~\cite{sanh2020movement, zafrir2021prune, xu2021rethinking}. The loss function is formulated as a linear combination of the standard loss associated with the specific task (e.g. cross-entropy for classification $\mathcal{L}_{CE}$) and the Kullback-Leibler divergence ($\mathcal{L}_{KL}$) between output distributions of the dense (teacher) model and the sparse (student) model in the form: $\mathcal{L}= (1-h) \mathcal{L}_{CE} + h \mathcal{L}_{KL}$. The ratio between the two is controlled with the \textit{hardness} hyperparameter $h$. To determine its optimal value at high sparsities we run an ablation study reported in Table \ref{tab:hardness_sweep}, and adopt the hardness $h=1$.

\begin{figure}[t]
    \centering
    \includegraphics[width=\linewidth]{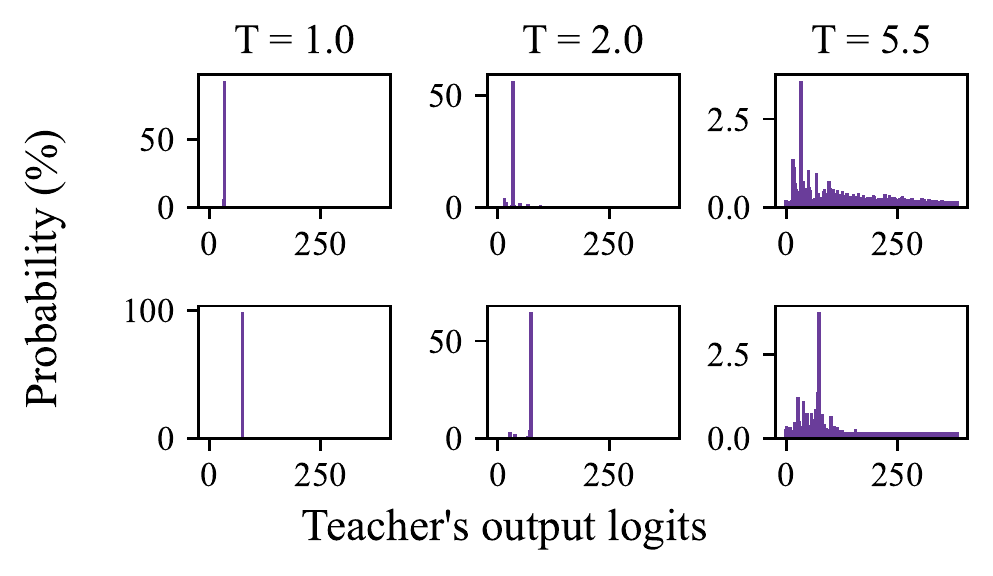}
    \vspace{-0.2in}
    \caption{Teacher's output distribution at commonly used temperatures ${T \in \{1.0, 2.0\}}$ and the proposed $T = 5.5$.}
    \label{fig:logits_dist}
    \vspace{-0.25in}
\end{figure}

\paragraph{Knowledge Distillation Temperature.} The temperature \textit{T} is an additional KD-hyperparameter that requires proper tuning, as it controls the ``softness'' of the output distribution. 
% \cite{hinton2015distilling} introduced it as a parameter to control the softness of the output distribution over logits in the form $q_i = \frac{exp(z_i/T)}{\sum_j exp(z_j/T)}$, where $q_i$ is a probability associated with the class $i$ based on its logit (score) $z_i$. Increasing $T$ leads to a softer distribution, which in the extreme case when $T \to +\infty$ leads to a uniform distribution over all classes. 
% The main intuition behind KD is to teach the sparse student to imitate its dense teacher. With temperature we try to make this task easier by rewarding the student even when its predictions are \textit{almost correct}. Contrary to this, the standard cross-entropy loss rewards the model for correct class predictions, which makes it an extremely hard task for sparse students. 
In the pruning literature, it is standard to use the ``stronger'' $T = 1 $ or $T = 2$ values \citep{xu2021rethinking, zafrir2021prune, sanh2020movement, lagunas2021block, kurtic2022optimal}; we revisit this by visualizing teacher's output distributions to get an insight into what the sparse student is learning. In Figure~\ref{fig:logits_dist}, we visualize generated distributions for randomly picked samples from the SQuADv1.1 task softened with three values of the temperature. As can be seen, teacher's high confidence in predicting the correct class at the commonly used temperatures $T \in \{1.0 , 2.0 \}$ makes the knowledge distillation almost obsolete. Motivated by this observation, we run an ablation study for many higher temperatures and report a fraction of results in Table~\ref{tab:temperature_sweep}. Given the results, we adopt the temperature $T = 5.5$.

\subsubsection{\gmp vs. other GMP-based baselines}
Due to space constraints, we aggregate all the previously analyzed improvements in a \textit{downstream pruning recipe} and present it in detail in Appendix~\ref{app:down_recipe}. We compare our optimized \gmp with other GMP results reported in the pruning literature. For a fair comparison, we consider both setups, 10 and 30-epoch. In the 10-epoch setup, we compare against the GMP baselines reported in \citet{sanh2020movement} and refer to them as \gmpmvp. In the 30-epoch setup, we compare against the best reported results in \citet{chen2020lottery}, obtained either via GMP or via Lottery Ticket (LTH) approach, and refer to them as \gmplth. As can be seen from the Table~\ref{tab:gmp_downstream}, our \gmp remarkably outperforms all other results; in some cases the improvements are more than \textbf{20 points}!

\begin{table}[t]
    \caption{Downstream pruning comparison of \gmp with other GMP-based baselines.}
    \label{tab:gmp_downstream}
    \centering
    \small{
      \begin{tabular}{lccccc}
        \toprule
        \multirow{2}{*}{Method} & \multirow{2}{*}{Spars.} & \multirow{2}{*}{Ep.} & SQuAD & MNLI & QQP \\
        & & & F1 & m-acc & acc \\
        \midrule 
        \bert & 0\% & & 88.5 & 84.5 & 91.1 \\
        \midrule
        \gmpmvp & \multirow{2}{*}{90\%} & \multirow{2}{*}{10} & 80.1 & 78.3 & 79.8 \\
        \gmp &  &  & \textbf{86.7} & \textbf{81.9} & \textbf{90.6} \\
        \midrule 
        \gmpmvp & \multirow{2}{*}{97\%} & \multirow{2}{*}{10} & 59.6 & 69.4 & 72.4 \\
        \gmp &  &  & \textbf{81.3} & \textbf{79.1} & \textbf{89.7} \\
        \midrule[1pt]
        \gmplth & \multirow{2}{*}{90\%} & \multirow{2}{*}{30} & 68.0 & 75.0 & 90.0 \\
        \gmp & & & \textbf{87.9} & \textbf{82.7} & \textbf{90.8} \\
        \midrule 
        \gmp & 97\% & 30 & 85.4 & 80.9 & 90.6 \\
        \bottomrule
      \end{tabular}
     }
      \vspace{-0.15in}
\end{table}

\subsubsection{\gmp vs. advanced pruning techniques}
Now, we wish to compare our \gmp with methods that rely on higher-order information to make pruning decisions, like gradients in MvP \cite{sanh2020movement} and the loss curvature in oBERT \cite{kurtic2022optimal}. Both of these impose higher computational overhead compared to the magnitude-based pruning, but we still put our results in the context with respect to theirs to fully grasp the scope of improvements introduced by careful optimizations of GMP. As can be seen from results in Table~\ref{tab:high_downstream}, \gmp is able to improve upon the performance of Movement Pruning in 4 out of 6 analyzed configurations, but unfortunately can't match the performance of the oBERT method. In addition to these comparisons, we make use of the open-source implementation of oBERT, current state-of-the-art BERT-pruning method, and run it with optimized hyperparameters from \gmp on the SQuADv1.1 task. We refer to these results as \obert. As can be seen from the Table \ref{tab:high_downstream}, even the very competitive oBERT results benefit from the \gmp setup. For all \gmp runs, we report mean performance across three runs with different seeds, and additional metrics in Tables \ref{tab:gmp_downstream2} and \ref{tab:high_downstream2}.

\begin{table}[t]
    \caption{Downstream pruning comparison of \gmp with advanced pruning techniques.}
    \label{tab:high_downstream}
    \centering
      \small{
      \begin{tabular}{lccccc}
        \toprule
        \multirow{2}{*}{Method} & \multirow{2}{*}{Spars.} & \multirow{2}{*}{Ep.} & SQuAD & MNLI & QQP \\
        & & & F1 & m-acc & acc \\
        \midrule 
        \bert & 0\% & & 88.5 & 84.5 & 91.1 \\
        \midrule
        \gmp & \multirow{2}{*}{90\%} & \multirow{2}{*}{10} & \textbf{86.7} & \textbf{81.9} & \textbf{90.6} \\
        MvP & & & 84.9 & 81.2 & 90.2 \\
        \midrule 
        \gmp & \multirow{2}{*}{97\%} & \multirow{2}{*}{10} & 81.3 & 79.1 & \textbf{89.7} \\
        MvP & & & \textbf{82.3} & \textbf{79.5} & 89.1 \\
        \midrule[1pt]
        \gmp & \multirow{3}{*}{90\%} & \multirow{3}{*}{30} & 87.9 & 82.7 & 90.8 \\
        oBERT & & & 88.3 & \textbf{83.8} & \textbf{91.4} \\
        oBERT$\scriptstyle\bigstar$ & & & \textbf{88.6} & & \\
        \midrule 
        \gmp & \multirow{3}{*}{97\%} & \multirow{3}{*}{30} & 85.4 & 80.9 & 90.6 \\
        oBERT & & & 86.0 & \textbf{81.8} & \textbf{90.9} \\
        oBERT$\scriptstyle\bigstar$ & & & \textbf{86.6} & & \\
        \bottomrule
      \end{tabular}
      \vspace{-0.15in}
  }
\end{table}
\subsection{Upstream pruning}

To validate the optimized \gmp setup introduced in the previous section, we apply it now to the pre-training phase of LLMs. This is a two-stage process. In the first stage, the \bert model is pruned during pre-training and then, in the second stage, the remaining weights are fine-tuned with the fixed mask on a specific downstream task to evaluate performance. Given the high costs of experimenting in the pre-training phase, we use the dense teacher open-sourced by~\citet{kurtic2022optimal}. Due to the space constraints, we summarize all hyperparameters in an \textit{upstream pruning recipe} and present it in detail in Appendix~\ref{app:up_recipe}. In Table \ref{tab:upstream} we present results obtained in this setup and compare against other methods that are utilizing the same approach. More specifically, we compare against the Lottery Ticket~\citep{chen2020lottery}, Prune OFA~\citep{zafrir2021prune}, and The Optimal BERT Surgeon (oBERT)~\citep{kurtic2022optimal}. In addition to this, we report the GMP baselines obtained in the Prune OFA work and refer to them as \gmpofa. As can be seen from the Table \ref{tab:upstream}, the \gmp significantly outperforms \gmpofa, Lottery Tickets and even the Prune OFA, and comes really close to the performance of oBERT. For all \gmp runs, we report mean performance across four runs with different seeds. These results confirm findings from the previous section and establish the \gmp as an extremely competitive baseline in all regimes.

\begin{table}
  \caption{Upstream pruning comparison of \gmp with other GMP-based baselines and more advanced pruning techniques.}
  \label{tab:upstream}
  \centering
  \small{
      \begin{tabular}{lcccc}
        \toprule
        \multirow{2}{*}{Method} & \multirow{2}{*}{Sparsity} & SQuAD & MNLI & QQP \\
        & & F1 & m-acc & acc \\
        \midrule 
        \bert & 0\% & 88.5 & 84.5 & 91.1 \\
        \midrule
        \gmpofa & 85\% & 86.2 & 82.5 & 90.9 \\
        \midrule
        Lottery Ticket & \multirow{4}{*}{90\%} & 68.0 & 75.0 & 90.0 \\
        Prune OFA & & 87.3 & 81.5 & 90.9 \\
        \gmp & & 88.2 & 83.2 & 90.8 \\
        oBERT & & \textbf{88.5} & \textbf{83.4} & \textbf{91.0} \\
        \midrule
        \gmp & \multirow{2}{*}{97\%} & 84.7 & 80.3 & 89.8 \\
        oBERT & & \textbf{84.9} & \textbf{80.9} & \textbf{90.3} \\
        \bottomrule
      \end{tabular}
      \vspace{-0.15in}
  }
\end{table}

\section{Conclusion}
In this work, we presented a set of updates to the standard gradual pruning setup for BERT models which enabled us to achieve very competitive results with the simple magnitude pruner. These results outperformed, by significant margins, all magnitude-based results currently available in the pruning literature which have been used as baselines for development and benchmarking of the new and more advanced pruning techniques. We hope that these \textit{new baselines} will help the community to start off from a competitive set of results when compressing large language models. Moreover, our \gmp has even outperformed some results obtained with more advanced and computationally heavier pruning techniques. At this point, we would like to {strongly emphasize} that these results should not be interpreted as evidence that magnitude pruning is better than other more advanced methods. Rather, they should be interpreted as evidence that their current results could significantly benefit from updates of the gradual setup presented on the \gmp use-case. To support this claim, we ran the state-of-the-art oBERT pruner with the \gmp setup and managed to improve its results by non-trivial margins.

\section{Limitations}
As any academic study, our work is not without its limitations. Following the literature, our extensive empirical studies were conducted only on the standard \bert model, giving us opportunity to compare against a vast amount of different pruning techniques. Throughout the literature, this model emerged as a consistent benchmark for unstructured pruning methods. However, the current results don't directly imply that our findings will be generally applicable to other language models as well. To partially fill in this uncertainty gap, we conduct a few experiments on the three times larger \bertL model and report results in the Appendix~\ref{app:additional_models}. Another limitation which we aim to remove in future work is the focus on fine-grained unstructured sparsity type, and explore other variants such as semi-structured and structured pruning. 

\bibliography{custom}
\bibliographystyle{natbib}

\appendix
\,
\section{Additional models}
\label{app:additional_models}
All of our experiments in the paper focus on the \bert model as it is the standard benchmark used in the pruning literature and we are able to compare results with a vast amount of other techniques. To verify that our proposed \gmp setup doesn't pertain only to the \bert model, in the Table~\ref{tab:bert_large} we present results on the three times larger $\textrm{BERT}_{\tiny{\textrm{LARGE}}}\,$ model. As a proof of concept, we run our downstream gradual pruning setup crafted for the \bert model without any hyper-parameter tuning. 

\begin{table}[!htb]
  \caption{Downstream pruning results when pruning the $\textrm{BERT}_{\tiny{\textrm{LARGE}}}\,$ model with the \gmp and gradual setup crafted for the \bert model.}
  \label{tab:bert_large}
  \centering
  \small{
      \begin{tabular}{lccc}
        \toprule
        \multirow{2}{*}{Method} & \multirow{2}{*}{Sparsity} & \multicolumn{2}{c}{SQuAD} \\
        & & F1 & EM \\
        \midrule 
        $\textrm{BERT}_{\tiny{\textrm{LARGE}}}\,$ & 0\% & 91.22 & 84.45 \\
        \midrule
        \gmp & 90\% & 90.12 & 83.21 \\
        \gmp & 97\% & 87.93 & 80.50 \\
        \bottomrule
      \end{tabular}
  }
\end{table}

\section{Downstream pruning recipe}
\label{app:down_recipe}
All of our implementations are built on top of HuggingFace's Transformers \footnote{https://github.com/huggingface/transformers} \cite{wolf-etal-2020-transformers} and Datasets \footnote{https://github.com/huggingface/datasets} \cite{lhoest-etal-2021-datasets} libraries, and NeuralMagic's SparseML \footnote{https://github.com/neuralmagic/sparseml} \cite{pmlr-v119-kurtz20a} library for model compression, and will be open-sourced to community along with our sparse models.

As our goal is to provide a simple and unique gradual pruning setup, all of our downstream runs (for all datasets) are using the same set of hyperparameters. The ones used to obtain results reported in Tables \ref{tab:gmp_downstream},  \ref{tab:high_downstream}, \ref{tab:gmp_downstream2}, and \ref{tab:high_downstream2} are as follows:
\begin{itemize}
    \item \texttt{learning-rate}: recurring 2-epoch scheduler (visualized in Figure \ref{fig:lr_and_spars}) with the initial value of \texttt{1e-4}, and the final value of \texttt{1e-6},
    \item \texttt{number-of-epochs}: 10 or 30 epochs, depending on the methods we compare against,
    \item \texttt{sparsity}: cubic scheduler with the initial pruning step of 70\% sparsity (visualized in Figure \ref{fig:lr_and_spars}),
    \item \texttt{pruning}: prune frequency of ten times per epoch, except during the first and last 2-epochs when only fine-tuning happens and masks are fixed,
    \item \texttt{student-initialization}: standard \bert (\texttt{bert-base-uncased}\footnote{https://huggingface.co/bert-base-uncased}),
    \item \texttt{knowledge-distillation (KD)}: (hardness, temperature) = (1.0, 5.5),
    \item \texttt{KD-teachers}: standard \bert fine-tuned on the corresponding task,
    \item \texttt{weight-decay}: 0.0,
    \item all other hyper-parameters are set to the standard default values, e.g.~\citet{sanh2020movement}:
        \begin{itemize}
            \item SQuADv1.1: \texttt{batch-size=16}, \texttt{max-sequence-length=384}, \texttt{doc-stride=128},
            \item MNLI and QQP: \texttt{batch-size=32}, \texttt{max-sequence-length=128}.
        \end{itemize}
\end{itemize}

\section{Upstream pruning recipe}
\label{app:up_recipe}
For a fair comparison with \citet{zafrir2021prune} and \citet{kurtic2022optimal}, we adopt the same gradual setup for pruning and fine-tuning, but apply our specific \gmp updates. The entire process is carried out in two stages. The first stage prunes the \bert model at upstream datasets, BookCorpus and English Wikipedia. Both datasets are available via~\citet{lhoest-etal-2021-datasets}. The \textit{upstream pruning recipe} can be summarized as follows:
\begin{itemize}
    \item \texttt{learning-rate}: recurring 0.5-epoch scheduler with the initial learning rate value of \texttt{5e-4}, and the final value of \texttt{5e-6},
    \item \texttt{number-of-epochs}: 3,
    \item \texttt{sparsity}: cubic scheduler with the initial pruning step of 70\% sparsity,
    \item \texttt{pruning}: prune frequency of hundred times per epoch, except during the last epoch when only fine-tuning happens and masks are fixed,
    \item \texttt{knowledge-distillation (KD)}: (hardness, temperature) = (1.0, 5.5),
    \item \texttt{KD teacher and student initialization}: \bert prepared and open-sourced by~\citet{kurtic2022optimal},
    \item \texttt{weight-decay}: 0.01,
    \item \texttt{batch-size}: 256,
    \item \texttt{max-sequence-length}: 512.
\end{itemize}
The second stage makes use of this upstream pruned model and fine-tunes it on a specific downstream task (SQuADv1.1, MNLI, QQP) for 8-epochs with fixed masks. All task-specific hyperparameters (batch-size, max-sequence-length, doc-stride, weight-decay) are the same as in Appendix~\ref{app:down_recipe}, and the remaining ones are as follows:
\begin{itemize}
    \item \texttt{learning-rate}: linear decay with initial value of \texttt{1.5e-5},
    \item \texttt{knowledge-distillation}: (hardness, temperature) = (1.0, 5.5),
    \item \texttt{KD-teachers}: standard \bert fine-tuned on the corresponding task.
\end{itemize}

\section{Additional metrics}
Due to space constraints, for corresponding runs in Tables \ref{tab:gmp_downstream}, \ref{tab:high_downstream}, and \ref{tab:upstream}, we present additional performance metrics in Tables \ref{tab:gmp_downstream2}, \ref{tab:high_downstream2}, and \ref{tab:upstream2}.

\begin{table}[h!]
    \caption{Downstream pruning comparison of \gmp with other GMP-based baselines. We report complementary evaluation metrics for results in Table~\ref{tab:gmp_downstream}.}
    \label{tab:gmp_downstream2}
    \small{
      \begin{tabular}{lccccc}
        \toprule
        \multirow{2}{*}{Method} & \multirow{2}{*}{Spars.} & \multirow{2}{*}{Ep.} & SQuAD & MNLI & QQP \\
        & & & EM & mm-acc & F1 \\
        \midrule 
        \bert & 0\% & & 81.4 & 85.0 & 88.0 \\
        \midrule
        \gmpmvp & \multirow{2}{*}{90\%} & \multirow{2}{*}{10} & 70.2 & 79.3 & 65.0 \\
        \gmp &  &  & \textbf{78.7} & \textbf{82.1} & \textbf{87.4} \\
        \midrule 
        \gmpmvp & \multirow{2}{*}{97\%} & \multirow{2}{*}{10} & 45.5 & 70.6 & 57.8 \\
        \gmp &  &  & \textbf{71.3} & \textbf{79.6} & \textbf{86.1} \\
        \midrule[1pt]
        \gmplth & \multirow{2}{*}{90\%} & \multirow{2}{*}{30} & - & - & - \\
        \gmp & & & \textbf{80.4} & \textbf{83.2} & \textbf{87.7} \\
        \midrule 
        \gmp & 97\% & 30 & 77.1 & 81.2 & 87.3 \\
        \bottomrule
      \end{tabular}
     }
\end{table}

\begin{table}[h!]
    \caption{Downstream pruning comparison of \gmp with advanced pruning techniques.  We report complementary evaluation metrics for results in Table~\ref{tab:high_downstream}.}
    \label{tab:high_downstream2}
    \centering
      \small{
      \begin{tabular}{lccccc}
        \toprule
        \multirow{2}{*}{Method} & \multirow{2}{*}{Spars.} & \multirow{2}{*}{Ep.} & SQuAD & MNLI & QQP \\
        & & &  EM & mm-acc & F1 \\
        \midrule 
        \bert & 0\% &  & 81.4 & 85.0 & 88.0 \\
        \midrule
        \gmp & \multirow{2}{*}{90\%} & \multirow{2}{*}{10} & \textbf{78.7} & \textbf{82.1} & \textbf{87.4} \\
        MvP & & & 76.6 & 81.8 & 86.8 \\
        \midrule 
        \gmp & \multirow{2}{*}{97\%} & \multirow{2}{*}{10} & 71.3 & 79.6 & \textbf{86.1} \\
        MvP & & & \textbf{72.7} & \textbf{80.1} & 85.5 \\
        \midrule[1pt]
        \gmp & \multirow{3}{*}{90\%} & \multirow{3}{*}{30} & 80.4 & 83.2 & 87.7 \\
        oBERT & & & 81.1 & \textbf{84.4} & \textbf{88.3} \\
        oBERT$\scriptstyle\bigstar$ & & & \textbf{88.6} & & \\
        \midrule 
        \gmp & \multirow{3}{*}{97\%} & \multirow{3}{*}{30} & 77.1 & 81.2 & 87.3 \\
        oBERT & & & 78.1 & \textbf{82.0} & \textbf{87.7} \\
        oBERT$\scriptstyle\bigstar$ & & & \textbf{78.8} & & \\
        \bottomrule
      \end{tabular}
  }
\end{table}

\begin{table}[h!]
  \caption{Upstream pruning comparison of \gmp with other GMP-based baselines and advanced pruning techniques.  We report complementary evaluation metrics for results in Table~\ref{tab:upstream}.}
  \label{tab:upstream2}
  \centering
  \small{
      \begin{tabular}{lcccc}
        \toprule
        \multirow{2}{*}{Method} & \multirow{2}{*}{Sparsity} & SQuAD & MNLI & QQP \\
        & & EM & mm-acc & F1 \\
        \midrule 
        \bert & 0\% & 81.4 & 85.0 & 88.0 \\
        \midrule
        \gmpofa & 85\% & 78.0 & 83.1 & 87.7 \\
        \midrule
        Lottery Ticket & \multirow{4}{*}{90\%} & - & - & - \\
        Prune OFA & & 79.8 & 82.4 & 87.7 \\
        \gmp & & 81.1 & \textbf{83.8} & 87.6 \\
        oBERT & & \textbf{81.4} & \textbf{83.8} & \textbf{87.8} \\
        \midrule
        \gmp & \multirow{2}{*}{97\%} & 76.3 & 81.0 & 86.5 \\
        oBERT & & \textbf{76.9} & \textbf{81.1} & \textbf{87.0} \\
        \bottomrule
      \end{tabular}
  }
\end{table}

\section{Ablation studies}

In Tables \ref{tab:initsparsity_sweep}, \ref{tab:lr_sweep}, \ref{tab:hardness_sweep}, \ref{tab:temperature_sweep} we present a subset of results from ablation studies conducted to find the optimal values of hyperparameters for the \gmp gradual pruning setup. These results illustrate the general trend of effects caused by varying one hyperparameter. Therefore, they don't cover all possible scenarios (i.e. higher-order effects when multiple hyperparameters are updated together), but such studies are computationally too expensive and we don't conduct them.

\begin{table}
        \caption{Initial sparsity step ablation study on the SQuADv1.1 dataset.}
        \label{tab:initsparsity_sweep}
        \centering
        \begin{tabular}{ccc}
            \toprule 
            \multirow{2}{*}{\makecell{Initial sparsity\\(\%)}} & \multicolumn{2}{c}{F1 score at} \\ 
            & 90\% & 97\% \\
            \midrule
            \phantom{0}0 & 85.2 & 77.2 \\
            30 & 85.5 & 77.8 \\
            50 & \textbf{85.8} & 78.5 \\
            70 & 85.8 & \textbf{79.1} \\
            \bottomrule
        \end{tabular}
\end{table}

\begin{table}
        \caption{Initial learning rate ablation study on the MNLI dataset.}
        \label{tab:lr_sweep}
        \centering
        \begin{tabular}{ccc}
            \toprule 
            \multirow{2}{*}{\makecell{Initial learning rate}} & \multicolumn{2}{c}{Accuracy at} \\
            & 90\% & 97\% \\
            \midrule
            3e-5 & 80.8 & 76.3 \\
            5e-5 & 81.4 & 77.8 \\
            8e-5 & \textbf{81.9} & 78.6 \\
            1e-4 & 81.6 & \textbf{79.3} \\
            \bottomrule
        \end{tabular}
\end{table}

\begin{table}
        \centering
        \caption{Knowledge Distillation (KD) hardness ablation study on the SQuADv1.1 dataset.}
        \label{tab:hardness_sweep}
        \begin{tabular}{ccc}
            \toprule 
            \multirow{2}{*}{\makecell{Knowledge distillation\\hardness}} & \multicolumn{2}{c}{F1 score at} \\
            & 90\% & 97\% \\
            \midrule
            0.6 & 84.6 & 78.4 \\
            0.8 & 85.9 & 80.1 \\
            0.9 & 86.2 & 80.7 \\
            1.0 & \textbf{86.7} & \textbf{81.0} \\
            \bottomrule
        \end{tabular}
\end{table}

\begin{table}
        \caption{Knowledge Distillation (KD) temperature ablation study on the SQuADv1.1 dataset.}
        \label{tab:temperature_sweep}
        \centering
        \begin{tabular}{ccc}
        \toprule 
        \multirow{2}{*}{\makecell{Knowledge distillation\\temperature}} & \multicolumn{2}{c}{F1 score at} \\
        & 90\% & 97\% \\
        \midrule
        1.0 & 84.7 & 77.3 \\
        2.0 & 85.8 & 79.0 \\
        5.5 & \textbf{86.7} & \textbf{81.0} \\
        8.5 & 86.4 & 80.9 \\
        \bottomrule
        \end{tabular}
\end{table}

\section{Learning rate schedulers we tried, but didn't work}
\label{app:failed_lr}
The schedulers we tried but didn't work: 1) linearly decaying learning rate, 2) the default fine-tuning learning rates (3e-5 for SQuADv1.1 and 2e-5 for MNLI and QQP), 3) learning rates with the warm-up phase. In the preliminary experiments, we have noticed that 1) and 2) have problems in recovering from the pruning steps at higher sparsities. The former one has extremely small learning rate values during the last few epochs when the model is pruned to high sparsities. The latter one continuously fails to recover properly even at moderate sparsity targets, which is why we run a sweep over a range of initial learning rate values. Given the results in Table \ref{tab:lr_sweep}, we decided to proceed with the 1e-4 as it helped to recover significantly at high sparsities. We haven't observed any benefits from the warmup phase, which is why we have decided not to use it as it adds an additional hyperparameter to tune.

\end{document}